# Device and System Level Design Considerations for Analog-Non-Volatile-Memory Based Neuromorphic Architectures

S. Burc Eryilmaz, Duygu Kuzum[+], Shimeng Yu[#], H.-S. Philip Wong*

Electrical Engineering Department, Stanford University, Stanford, CA 94305, [+]Electrical and Computer Engineering Department, University of California, San Diego, [#]Arizona State University, *e-mail: hspwong@stanford.edu

**Abstract**
This paper gives an overview of recent progress in the brain-inspired computing field with a focus on implementation using emerging memories as electronic synapses. Design considerations and challenges such as requirements and design targets on multilevel states, device variability, programming energy, array-level connectivity, fan-in/fan-out, wire energy, and IR drop are presented. Wires are increasingly important in design decisions, especially for large systems; and cycle-to-cycle variations have large impact on learning performance.

### Brain-inspired Algorithms

There are two broad classes of brain-inspired algorithms: (a) biology-based learning models (such as spike-timing-dependent-plasticity (STDP)) and neuron dynamics [1-3] that are models of neural populations developed from studying the brain, and (b) artificial neural networks (ANNs) adapted to solve machine learning (ML) tasks, and these do not strictly mimic the biology, but are inspired by the brain to some extent [4,5] while being optimized for hardware implementation. In order to scale up both classes of brain-inspired algorithms in system size within power/energy constraints, hardware customization is critical [6].

***Conventional hardware for brain-inspired computing:*** Conventional hardware architectures, such as CPUs, GPUs, and supercomputers [7-11] have been used for brain-inspired computing. With 100s of kWs of power consumed, large-scale implementations are not accessible to researchers except the few with huge amount of computing power (see Figs. 1, 2).

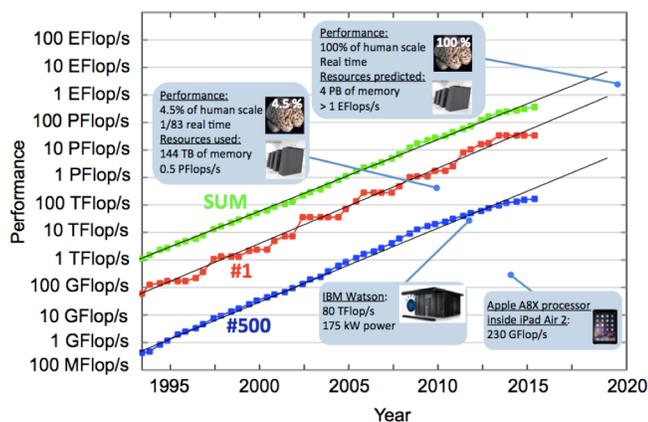

Figure 1. Performance improvements of supercomputers with years, compared with different types of hardware. Adapted from [7, 12].

***Brain-inspired hardware***: The development of neuromorphic hardware [13-15] has started decades ago (see Fig. 3). It employs connectivity, processing, and communication schemes similar to those of the brain on the device, circuit [13-16], and the architecture level [17-20]. It has been used to simulate the biological brain for scientific purposes [8, 14], and perform practical learning/inference tasks [16, 19]. By analogy to the biological brain, summing nodes in neuromorphic hardware and the connections between them that store the weights are referred as neurons and synapses, respectively.

| Application | | Hardware used | Total time for the experiment | Estimated power consumption |
|---|---|---|---|---|
| Large scale | Emulating 4.5% of human brain: $10^9$ neurons, $10^{13}$ synapses [7] | Blue Gene/P: 36,864 nodes, 147,456 cores | 1/83[th] real time of brain | **2.9 MW** with LINPACK benchmark [45] |
| | Training a deep sparse autoencoder – $10^9$ synaptic weights trained with 10M images 200x200 pixels each [8] | 1,000 CPUs (16,000 cores) | 3 days | **~100 kW** (cores only, does not include power consumption of memory or network) |
| Small to moderate scale | Convolutional neural net with 60M synaptic weights and 650K neurons trained with 1.2M images 256x256 pixels each [9] | 2 Nvidia GTX580 3GB GPUs | 5-6 days | **1,200 W** |
| | Restricted Boltzmann Machine trained by contrastive divergence: 28 M synaptic weights trained in a network with 20,736 visible and 49,152 hidden neurons with 1M images 144x144 pixels each [10] | Nvidia GeForce GTX280 GPU | 3,415 s | **550 W** |
| | | Dual Core CPU at 3.16 GHz | 32,236 s | **65 W** |
| | Processing 1 s of speech using deep neural network [11] | Nvidia Tesla C2070 GPU | 20-490 ms | **238 W** |
| | | Intel Xeon DP Quad Core E5640 | 1360 ms | **80 W** |

Figure 2. Type of conventional hardware and their power consumptions used for several brain-inspired computing tasks. For the first row (BlueGene), data is obtained from [45]. For other rows, power consumption values are thermal design power (TDP) from the manufacturer's website.

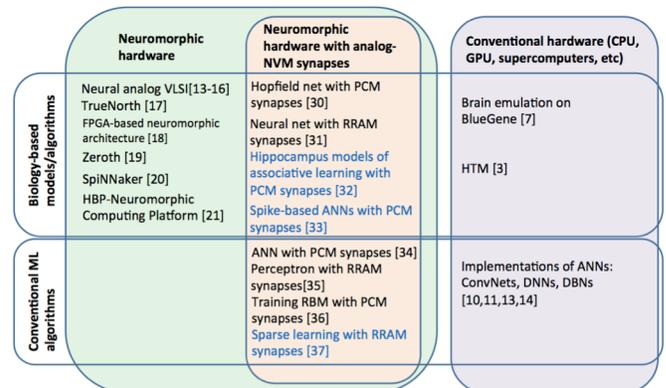

Figure 3. Taxonomy for algorithms and hardware technologies for brain-inspired computing. Brain-inspired algorithms include conventional machine learning (ML) algorithms and biology-based models (based on spiking neurons, STDP learning, etc.). Neuromorphic hardware is being developed in addition to the use of conventional hardware such as supercomputers. Neuromorphic hardware is a non-conventional hardware paradigm that employs connectivity, processing, and communication schemes similar to the brain. Analog-non-volatile-memory (a-NVM) based neuromorphic hardware uses a-NVM for emulating synapses in a brain-inspired fashion. Works highlighted with blue are based on simulation works that use single device models obtained from experimental data. Acronyms: RBM- restricted Boltzmann machine, HTM- hierarchical temporal memory; ConvNet-convolutional neural net; DNN-deep neural net; DBN-deep belief net, ANN-artificial neural network.

*Neurons*: Neurons can be digital [46] or analog [15], and may communicate asynchronously [17]. IBM's all-digital design TrueNorth [17] is energy-efficient due to the locality of the synapses (SRAM) with the neurons. The SRAM synapse is volatile and expensive in terms of area (>120 $F^2$) and has limitations in terms of scaling up in system size.

*Synapses:* Because the number of synapses in a neural network far exceeds the number of neurons, the power, performance, device density, and wiring of the electronic synapses deserve special attention. "New" non-volatile resistive memory elements such as resistive switching memory (RRAM) [22], phase change memory (PCM) [23], conductive bridge memory (CBRAM) [24], and ferroelectric memory (FeRAM) [25] have characteristics that are desirable as electronic synapses (spin-transfer-torque memory (STT-MRAM) has low dynamic range and is not suitable). These are two terminal devices with excellent size scalability, low energy operation, analog programmability [26, 27]. Monolithic 3-dimensional integration of NVM with CMOS was demonstrated [28,29], which allows the designers to hide the neuron circuitry underneath multiple layers of synapses (see Fig. 4). A variety of algorithms have utilized their gradual resistance change as a synapse (Fig. 5) [30-37]. These devices can also implement various variations of the STDP rule (Fig. 6), further suggesting their use for neuromorphic hardware [26].

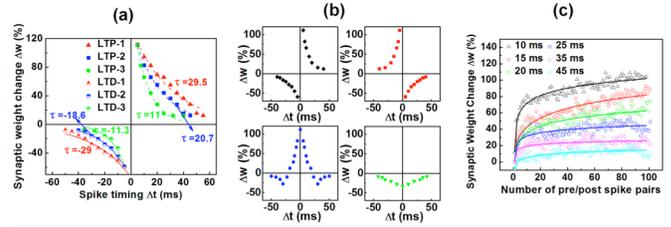

Figure 6. A variety of STDP kernels and STDP behavior can be implemented with the PCM device: (a) Various time constant (b) Different STDP kernels (c) Saturation of synaptic weight change in PCM synapse [23]

**Design Considerations and Open Problems**

*Electronic synapse device*: $R_{off}/R_{on} = 10^3 – 10^4$ are shown to give good results in simulations for unsupervised learning tasks [37, 38], but these studies ignore many non-idealities of the RRAM device such as the need for a realistic non-linear behavior as well as inherent device-to-device and cycle-to-cycle variations. For digital synapse, simulations show that 5- or 6-bit (32-64 levels) weights are acceptable for some ML tasks [37, 39]. Using 5-bit synapse reduces the recognition accuracy only slightly to 89.4% compared to 91.9% obtained with 64-bit synapse [39], although this analysis uses 64-bit (double precision) synapses in training and uses 5 bits only at the time of inference. While 100 or more levels have been observed with new NVM devices [23, 27], conductance change is not uniform over the full conductance range and precise setting of conductance level is often impossible using single-shot programming [23, 27]. Hence, effects of non-uniform/imprecise conductance change should be mitigated on the algorithm and task level while pursuing device engineering for better control over the gradual conductance change. Some devices show gradual resistance change only in one direction (SET or RESET). To address this shortcoming, stochastic switching was proposed for RRAM and CBRAM [40, 41], but it requires re-formulation of the algorithms that are based on multi-level synapses. Alternatively, the synaptic weight can be stored in two devices with a differential read-out to make both directions artificially gradual (Fig. 7) [33], trading off synapse density. When a realistic RRAM model calibrated to measurements over a number of devices is employed [42] for 2-RRAM synapses, we have found that $R_{off}/R_{on}$ ratio of 500 results in 88 % recognition accuracy (out-of-sample) on MNIST dataset for a supervised contrastive divergence training of a 2-layer restricted Boltzmann machine (RBM). This is lower than state-of-the-art (99.79%) [43], but close to what a similar RBM achieves with 64-bit digital synapses (92%) [39]. We observe that using a differential read-out with 2-RRAM synapses relaxes $R_{off}/R_{on}$ requirement on memory cells, since training can bring the conductances of the two cells in a synapse very close to each other to realize a 0 weight, even if $R_{off}/R_{on}$ is less than $10^3$. This study initializes RRAM cells in the low-R state and performs gradual RESET during training, using a refresh mechanism shown in Fig. 7. For NVM devices, cycle-to-cycle and device-to-device variations exist. We observe that when 2-RRAM synapse is employed, device-to-device variations are tolerable to a big extent since training reduces the effect of device-to-device variations in a way to make the differential conductance minimize the overall error. In fact, learning is much more efficient when initial weights are randomly initialized for symmetry breaking instead of initializing all to a same value [44]. Initializing random weights requires pseudo-random number generator hardware on conventional computers, whereas it is easily done on NVM arrays through their intrinsic device-to-device variations. On the other hand, learning performance is a strong function of number of gradual levels for the NVM and cycle-to-cycle variation. Fig. 8 shows how classification error of a network trained on MNIST dataset is affected by these two. While training is robust to variations to some extent, there is a point where classification error starts to get worse. The classification error is a function of cycle-to-cycle variation as well as average conductance change. It is important to note that learning performance can also be affected by nonlinearity when nonlinearity is large. Such analysis that takes into account the effect of variations and nonlinearity should be performed for algorithms that are aimed to be implemented on analog-NVM based hardware, which emphasizes the importance of the interplay between algorithm and device behavior. Device models that can accurately capture temporal variations as a function of conductance and applied voltage are essential.

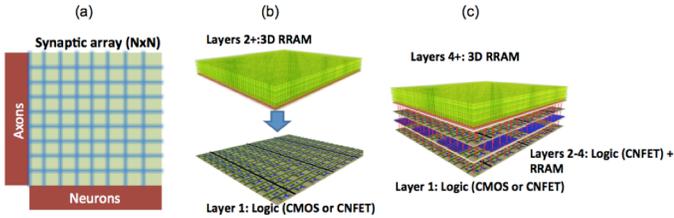

Figure 4. Compared to 2D crossbar implementation (a), 3D stacking hides the neuron circuitry by integrating synapses on top of neurons (b). New technologies such as carbon nanotube transistors can also enable logic/memory hybrids [28] on different layers as well as within the same layer (c).

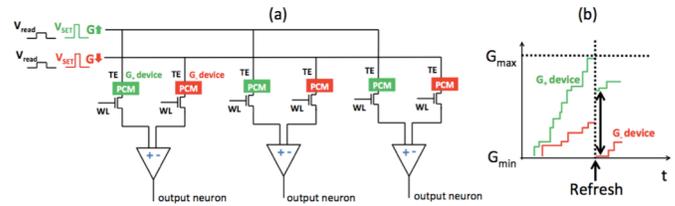

Figure 7. 2-cell synapse scheme (a) and refresh mechanism (b) [33].

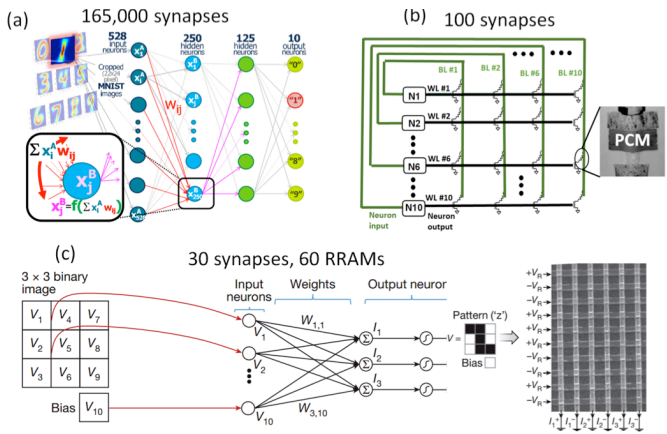

Figure 5. Experimental demonstrations of small and mid-scale networks using resistive memory devices: (a) Digit recognition with a multilayer feedforward network with 165,000 synapses [34] (b) Pattern recognition in a Hopfield network with 100 synapses [30], (c) Pattern recognition with 30 synapses and 60 RRAM devices (2 RRAM per synapse) [31]. For all these demonstrations, synapses are on hardware (memory devices) and neuron functionality is on software.

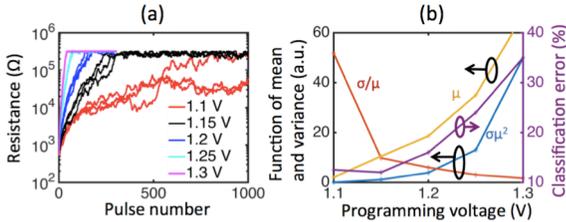

Figure 8. Gradual resistance change in RRAM device for different programming voltages (a) and how cycle-to-cycle variations as well as gradual conductance change affect learning performance (b). μ refers to average log-conductance change for one pulse over a cycle and σ refers to standard variation of fluctuations of log-conductance change around a smooth fit over a cycle. Number of gradual levels obtained for 1.1, 1.15, 1.2, 1.25 and 1.3 V are ~1000, 270, 170, 90, 45 respectively, and increases with decreased μ for a given $R_{on}/R_{off}$. Classification accuracy is a function of both number of gradual levels (or, equivalently, μ) and cycle-to-cycle variation, as seen in (b).

There are cases where too much device-to-device variations can hurt performance but can be ameliorated with increased latency; these cases must be quantified in terms of the energy and latency tradeoffs [47]. Experiments using a Hopfield net shows that an increase in device-to-device variability from 9% to 60% can be tolerated with an increase in energy consumption in PCM synapses from 4 to 54 nJ, and using 11 iterations instead of 1 [47]. On edge detection and digit recognition tasks, 25% variability in RRAM synapses results in negligible degradation in the output [27, 38]. Using redundant cells can reduce the variation, trading off area and energy [48]. Programming energy of these devices is also crucial. RRAM can go down to a few ns of write time and <pJ of programming energy (Fig. 9a), and is a good candidate as a synaptic device if variability can be addressed. PCM is also promising because it has less resistance variation, but integration on smaller technology nodes is needed to reduce the write energy [26]. The longer write times (>10 ns) of PCM might not be a key limiting factor due to the highly parallel nature of brain-inspired hardware (see Fig. 9a).

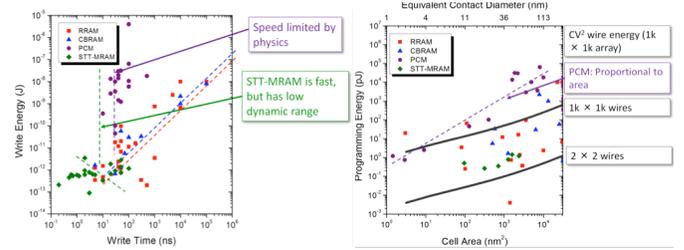

Figure 9. (a) Write time vs write energy for several emerging memory technologies. (b) Programming energy vs cell area superimposed with wire energy for several memory technologies. The wire energy is computed for a 1k × 1k crossbar array: top line is the energy for energizing all 1k × 1k (two thousand) wires, bottom line is the energy consumed by energizing 2 × 2 (four) wires. Data source: [49]

*Connectivity and Fan-Out*: Brain-inspired computing algorithms require neurons with large fan-in and fan-out. Energy consumption in synapse arrays have two major components: wire energy ($CV^2$) and programming energy ($V^2 t_{pulsewidth}/R$). Wire energy considerations are important because even for a 1k×1k array, wire energy (order of 1 pJ) starts dominating (see Fig. 9b, 10). Low programming voltage (potentially offered by CBRAM) is strongly desired as it lowers both programming and wire energy, which decreases with $V^2$. To further illustrate energy considerations, we perform energy consumption analysis in our case study of RBM training for MNIST digit recognition. When low-R of RRAM cells is ~600 Ω, RRAM energy is 3 orders of magnitude higher than wire energy. RRAM energy roughly reduces linearly with reduced conductance, and it is comparable to wire energy when low-R is 600 kΩ. Fig. 10 illustrates this case, and provides a guideline for tradeoffs related to energy consumption. Wire energy directly scales quadratically with the programming voltage. On the other hand, RRAM energy has more complex dependencies on programming voltage. When RRAM cells are initialized in low-R states before training, if training takes only a few epochs, lowering programming voltage too much results in RRAM cells spending more time in low-R states, which significantly increases energy consumption. This effect fades away after several refresh cycles (see Fig. 7), and almost disappears after 50 epochs. In our case, lowest classification error was achieved in 3 epochs, hence reducing the programming voltage to 1.1 V results in unnecessary energy consumption in RRAM cells, which is important unless wire energy is the dominating factor. If training continues for more than 50 epochs, programming energy scales directly with $V^2$.

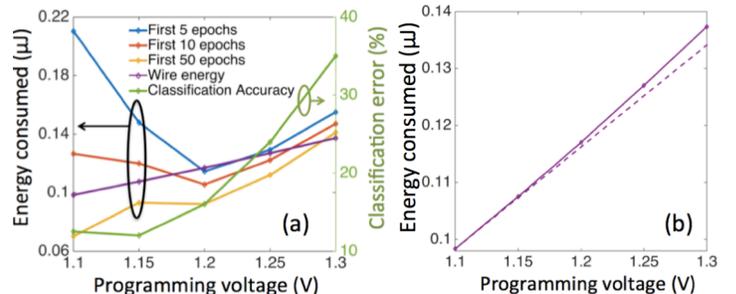

Figure 10. (a) Energy consumption (training phase) vs. classification error after training. Blue, red and orange curves are energy consumed per epoch (averaged over first 5, 10 and 50 epochs) within RRAM for a low-R value of 600 kΩ. Wire energy assumes 1k×1k array with 100 nm full pitch. (b) Inset shows the quadratic dependence of wire energy (solid curve) by comparing it with a line (dashed).

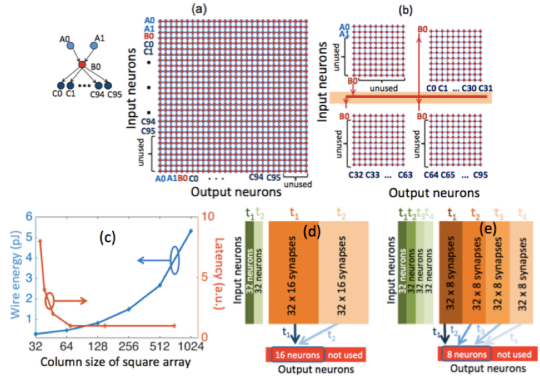

Figure 11. (a) Mapping a neural network architecture on one big array (128×128) vs (b) several smaller 32×32 arrays. In this scenario, second approach saves 60 % wire energy to the first order. (c) Energy savings reduce due to communication overhead as smaller arrays are used. Time-multiplexing can be employed to obtain larger fan-in in small arrays ((d) and (e)), but this introduces latency (c). Analysis in (c) is done for a 256-4-256 autoencoder network. See [50] for autoencoder reference.

Using a huge array to accommodate the largest fan-out in a network is energy inefficient due to charging and discharging of long wires within the array. If the connectivity of the network or neuron activations are sparse, using multiple small corelets instead of big arrays reduces wire energy, by 60% for the example in Fig. 11a,b when four 32×32 arrays are used instead of one 128×128 array. However, a first order analysis of communication cost between corelets in terms of wire energy shows that energy benefits of using smaller arrays reduce due to communication overhead, using AER (address event representation) as an example communication means (Fig. 11c). Furthermore, smaller arrays might require time-multiplexing of input neurons to accommodate large fan-ins (Fig. 11d, e), which introduces latency as shown in Fig. 11c. The tradeoffs between latency, energy savings and circuit complexity should be considered when choosing the array size.

IR drop along the wires is an important issue when choosing the array size. To mitigate the effects of IR drop, RRAM cells should be operated in higher R regimes if larger arrays are desired (Fig. 12). Larger arrays of memory cells with low $R_{on}$ are also more prone to read inaccuracy, due to IR drop along the wires. When an ADC is employed to convert analog current to digital data, increasing $R_{on}$ from 10 kΩ to 1 MΩ reduces the read inaccuracy from 20% to less than 1%, while reducing the read energy from 1 pJ to 20 fJ [37]. Reducing IR drop by increasing wire width can reduce read inaccuracy [37] but this consumes more area and introduces larger wire capacitances. The factors that determine the array size are: connectivity of network architecture, $R_{on}$ and $R_{off}/R_{on}$ ratio, as well as the area of neuron vs synapse.

## Conclusion

Emerging NVM devices have characteristics that are superior to conventional alternatives for emulating synaptic functionality and enabling on-line learning; but require careful engineering of device properties and array architecture to suit the network topology and algorithm. Wires are as important as the devices and innovations in scalable connectivity will be beneficial for large-scale systems.

## Acknowledgments

This work is supported in part by SONIC, one of six centers of STARnet, a Semiconductor Research Corporation program sponsored by MARCO and DARPA, the NSF Expedition on Computing (Visual Cortex on Silicon, award 1317470), and the member companies of the Stanford Non-Volatile Memory Technology Research Initiative (NMTRI) and the Stanford SystemX Alliance. Collaborations and discussions with researchers at IBM and Siddharth Joshi, Gert Cauwenberghs and Emre Neftci (UCSD) are highly appreciated.

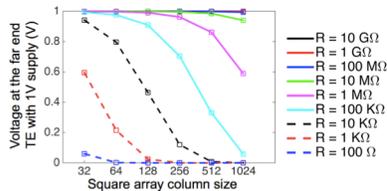

Figure 12. Voltage at top electrode of the RRAM cell on the far side from the 1 V supply, for several array sizes and RRAM resistances.